\pdfoutput=1

\documentclass[11pt]{article}

\usepackage[final]{acl}

\usepackage{times}
\usepackage{latexsym}
\usepackage{amsfonts}

\usepackage[T1]{fontenc}

\usepackage[utf8]{inputenc}

\usepackage{microtype}
\usepackage{amsmath}

\usepackage{booktabs}
\usepackage{algorithm}
\usepackage{algorithmic}
\usepackage{graphicx}
\usepackage{url}
\usepackage{makecell}

\title{AutoLoRA:  Automatically Tuning Matrix Ranks in Low-Rank Adaptation Based on Meta Learning}

\author{Ruiyi Zhang$^{1*}$,  Rushi Qiang$^{2*}$,  Sai Ashish Somayajula$^1$, Pengtao Xie$^1$ \\
  $^1$UC San Diego $^2$Tsinghua University\\
  \texttt{\{ruz048, p1xie\}@ucsd.edu}
}

\begin{document}
\maketitle

\newcommand\blfootnote[1]{%
  \begingroup
  \renewcommand\thefootnote{}\footnote{#1}%
  \addtocounter{footnote}{-1}%
  \endgroup
}

\blfootnote{*Equal contribution.}

\begin{abstract}
Large-scale pretraining followed by task-specific finetuning has achieved great success in various NLP tasks. Since finetuning all parameters of large pretrained models poses substantial computational and memory challenges, several efficient finetuning methods have been developed. Among them, low-rank adaptation (LoRA), which finetunes low-rank incremental update matrices on top of frozen pretrained weights, has proven particularly effective. Nonetheless, LoRA's uniform rank assignment across all layers, along with its reliance on an exhaustive search to find the best rank, leads to high computation costs and suboptimal finetuning performance. To address these limitations, we introduce AutoLoRA, a meta learning based framework for automatically identifying the optimal rank of each LoRA layer. AutoLoRA associates each rank-1 matrix in a low-rank update matrix with a selection variable, which determines whether the rank-1 matrix should be discarded. A meta learning based method is developed to learn these selection variables. The optimal rank is determined by thresholding the values of these variables. Our comprehensive experiments on natural language understanding, generation, and sequence labeling demonstrate the effectiveness of AutoLoRA. The code is publicly available at \url{https://anonymous.4open.science/r/AutoLoRA}. 

\end{abstract}

\section{Introduction}
\label{sec:intro}

Large Language Models (LLMs)~\cite{gpt2,gpt3} have demonstrated state-of-the-art performance across a variety of NLP tasks, spanning from Natural Language Understanding (NLU)~\cite{glue} to Natural Language Generation (NLG)~\cite{dart}, a trajectory highlighted by the success of models like ChatGPT~\cite{openai2023gpt4}. 
Their success largely stems from a two-stage process: initial pretraining on vast amounts of unlabeled texts, followed by finetuning on specific downstream tasks. However, as models scale up, for instance transitioning from RoBERTa-large's 355 million parameters~\cite{RoBERTa} to GPT-3's staggering 175 billion parameters~\cite{gpt3}, finetuning becomes highly expensive in computation. 

To address this challenge, many efficient finetuning methods~\cite{adapter} have been developed. 
For instance, the Adapters method~\cite{adapter} inserts   lightweight layers (called adapters) into pretrained networks. During finetuning, only these adapters are updated while the pretrained layers are kept frozen. One limitation of this method is that the adapters incur additional computation overhead during inference. Another approach, prefix tuning~\cite{prompt}, introduces trainable prefix parameters which are prepended to the input sequence while making the pretrained model parameters frozen. Nevertheless, determining the optimal length of the prefix can be tricky. A prefix that is too short cannot capture enough information, while an overlong prefix may largely reduce the maximum length of the input sequence. To address these limitations, LoRA~\cite{lora} proposes to add low-rank incremental update matrices to pretrained weight matrices. During finetuning, only the incremental matrices are trained while the pretrained ones are frozen. The low-rank parameterization significantly reduces the number of finetuning parameters.

\begin{figure*}
    \centering
    \includegraphics[width=0.9\textwidth]{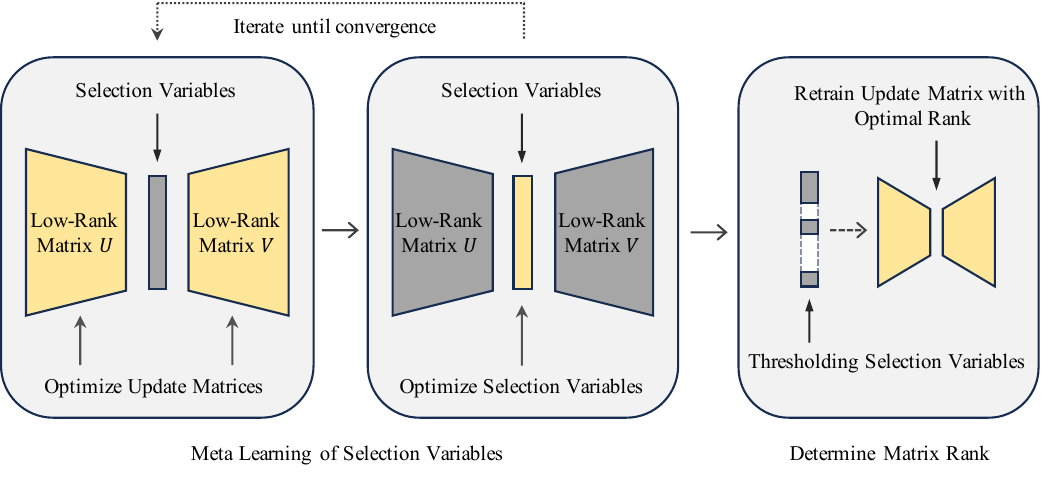}
    \caption{An overview of AutoLoRA. In the meta learning process, AutoLoRA learns selection variables with two iterative steps. Firstly, the weights in the update matrices are optimized on the training dataset. Secondly, the selection variables are updated on the validation dataset. These two steps are iterated until convergence is achieved. Upon acquiring the optimal values of the selection variables, AutoLoRA determines the optimal matrix ranks by thresholding these values. Subsequently, the ranks of update matrices in the LoRA layers are set to the learned optimal ranks and 
    retrained on the combination of training and validation data.}
    \label{fig:1}
    \vspace{-0.4cm}
\end{figure*}



While achieving parameter-efficient finetuning without increasing inference costs, LoRA has two limitations. First, the update matrices at different  layers share the same rank, without considering the varying properties across layers. Different layers in a pretrained model 
have varying importance to a downstream task and should be adapted differently, which requires the number of trainable parameters to be layer-specific.  Employing a uniform rank across all layers compromises this purpose, which renders   some layers to be under-parameterized (leading to suboptimal finetuning  performance) while others unnecessarily over-parameterized (leading to  computation inefficiency).  
Second, obtaining the optimal rank in LoRA typically involves an extensive manual hyperparameter search, which is time-consuming and poses scalability issues.

To address the aforementioned limitations of LoRA, we introduce the AutoLoRA framework to automatically determine the optimal rank for each LoRA layer. In AutoLoRA, we first decompose an  update matrix into the product of two low-rank  matrices (with rank $k$), in alignment with the LoRA methodology. This product can be expressed as the summation of $k$ rank-1 matrices. For each  rank-1 matrix, we assign a continuous  trainable  selection variable $\alpha\in[0,1]$  indicating the matrix's relative importance in the summation. After learning, if $\alpha$ is close to zero, the corresponding rank-1 matrix is removed from the summation. These selection variables effectively control the rank of an update matrix. Learning $\alpha$ directly on a training dataset together with the update matrices can result in overfitting, and the network learned in this way lacks generalization ability. To mitigate this problem, we formulate the search process of $\alpha$ as a meta learning~\cite{maml} problem.  First, we finetune the weights in the rank-1 matrices on a  training dataset. Second, we optimize the $\alpha$ values by minimizing the loss on a validation dataset. These two steps iterate until convergence. 
Subsequently, we derive the optimal rank of each LoRA layer by thresholding the learned $\alpha$ values. Once the optimal rank is  identified for each layer, the weights in  the low-rank update matrices are retrained on the combination of training and validation data. An overview of our proposed method is illustrated in Figure \ref{fig:1}.


The major contributions of this paper are summarized as follows. 
\begin{itemize}
    \item We propose AutoLoRA, a meta learning based approach that can automatically determine the optimal and layer-specific ranks of update matrices,  
    alleviating the burden of manually tuning them as in LoRA. 
    \item 
    Extensive experiments on 
    natural language understanding and generation tasks demonstrate the effectiveness of AutoLoRA.
\end{itemize}

\section{Related Works}

\subsection{Parameter Efficient Finetuning Methods
}

Various methods have been developed for efficiently finetuning  pretrained models. These methods update only a small subset of the weights in large pretrained models, leaving the majority of the parameters frozen. According to \citet{intrinsic}, weight matrices in large pretrained models tend to have a small intrinsic dimension, offering theoretical intuitions for finetuning pretrained models with low-dimensional reparameterization. Impressively, these methods can sometimes surpass the performance of full finetuning, particularly in downstream tasks with limited training data. 

Some efficient finetuning methods finetune the pretrained model by updating trainable prompts while leaving its pretrained parameters frozen. For example, Prompt-tuning \cite{prompt} learns ``soft prompts'' for language models to perform specific downstream tasks. Prefix-tuning \cite{prefix} optimizes a sequence of continuous task-specific vectors for natural language generation tasks.  P-tuning \cite{ptuning} optimizes a small neural network which generates continuous prompt embeddings to finetune GPT models for natural language understanding tasks. LLaMA-Adapter \cite{llamaadapter} learns trainable prompts for the LLaMA~\cite{llama} model. However, selecting appropriate prompt length can be challenging, as short prompts cannot capture sufficient information while overlong prompts significantly reduce the input sequences' length.

Another line of research involves finetuning the pretrained model by inserting trainable modules into the model while keeping pretrained parameters frozen. For example, Adapter \cite{adapter} proposes to inject additional trainable adapter layers into pretrained Transformer~\cite{transformer} models.  IA3 \cite{ia3} multiplies the output of activation functions in the pretrained model with trainable vectors. Compacter~\cite{compacter} inserts hypercomplex multiplication layers \cite{hypercomplex} to the pretrained model, offering more efficiency than those in Adapters. These methods incur additional inference overhead due to computing the inserted modules. 

AdaLoRA~\cite{adalora} aims to overcome the problem that LoRA evenly distributes the budget of updates across all LoRA layers by adaptively allocating the budget according to their importance scores. However, since both the importance score and update matrices are learned on the same training dataset, there is an increased risk of overfitting.

\subsection{Meta Learning}

Various meta learning methods have been proposed for better adaptation of models to new tasks with minimal training data. For instance, Model-Agnostic Meta-Learning (MAML)~\cite{maml} is a gradient based meta learning method, aiming to train model weights for fast adaptation to new tasks with small amounts of data in a few gradient descent steps. Meta-SGD is an extension  of MAML~\cite{metasgd}. It not only learns  model weights, but also optimizes  learning rates for fast adaptation to new tasks. Reptile~\cite{reptile} is a first-order meta learning algorithm, which serves as a simpler alternative to MAML. Reptile repeatedly moves the initialization of meta parameters towards the model weights trained on a specific task, sidestepping  second-order gradient computation.  Orthogonal to these previous methods, our meta learning based method is used for tuning matrix ranks in LoRA.

\section{Preliminaries}


In LoRA \cite{lora}, a weight matrix $W_l\in \mathbb{R}^{m_l\times n_l}$ at layer $l$ in a  downstream model is parameterized as $W_l=\widetilde{W}_l+\Delta_l$, where $\widetilde{W}_l$ is the weight matrix at layer $l$ in a pretrained model and $\Delta_l$ is an incremental update matrix.  $\Delta_l$  is parameterized as the product of two low-rank matrices: $\Delta_l= U_lV_l$, where $U_l \in \mathbb{R}^{m_l\times k_l}$ and $V_l \in \mathbb{R}^{k_l\times n_l}$. $k_l$, which is much smaller than $m_l$ and $n_l$, is the rank of $\Delta_l$. Equivalently, $\Delta_l$ can be written as the summation of $k_l$ rank-1 matrices:
\begin{equation}
\label{eq:rank1}
    \Delta_l=\sum_{j=1}^{k_l}\Delta_l^j, 
\end{equation}
where $\Delta_l^j$ is the outer-product between the $j$-th column of $U_l$ and the $j$-th row of $V_l$.

\begin{algorithm*}[t]
\caption{AutoLoRA: Automatically  Search for Optimal Rank}\label{alg:1}
\begin{algorithmic}
\STATE Initialize selection variables $A=\{\alpha_l\}_{1\leq l\leq M}$ and  update matrices  $\{\Delta_l=\sum_{j=1}^{k_l}\alpha_l^j\Delta_l^j\}_{1\leq l\leq M}$
    \WHILE{\textit{not converged}}
        \STATE 1. Update weight parameters $W$ by descending $\nabla_W\mathcal{L}_{tr}(W,D_{tr})$.
        \STATE 2. Update selection variables $A$ by descending $\nabla_A\mathcal{L}_{val}(W-\eta\nabla_W\mathcal{L}_{tr} (W,D_{tr}),D_{val})$. \\
    \ENDWHILE
\STATE Derive the best rank $k^*_l=|\{\alpha_{l,j}^{*}|1\leq j\leq k_l, \alpha_{l,j}^{*}\geq \lambda\}|$ for LoRA layer $l$ from optimal values  $\alpha^*_l$. 
\end{algorithmic}
\end{algorithm*}

\section{Method}

\subsection{Overview}
In AutoLoRA, we aim to automatically determine the rank $k_l$ in Eq.(\ref{eq:rank1}), instead of manually specifying it as in LoRA. To achieve this goal, we associate each rank-1 matrix in an update matrix with a selection variable and reparameterize the update matrix as a weighted sum of rank-1 matrices. A meta learning based approach is developed to learn these selection variables. After learning, if the value of a selection variable is close to zero, its corresponding rank-1 matrix is removed. In this way, we can determine the optimal rank for each update matrix based on the selection variables. An overview of the AutoLoRA algorithm is shown in Algorithm \ref{alg:1}.

\subsection{Reparameterize  Update Matrices}
\label{sec:rep}
We associate each rank-1 matrix $\Delta_l^j$ in Eq.(\ref{eq:rank1}) with a selection variable $\alpha^j_l\in[0,1]$ and reparameterize $\Delta_l$ as  a  weighted sum of rank-1 matrices: 
\begin{equation}
\Delta_l=\sum_{j=1}^{k_l}\alpha_l^j\Delta_l^j.
\end{equation} 
$\alpha^j_l$ can be interpreted as the importance of $\Delta_l^j$.  
If $\alpha^j_l$ is close to 0,  $\Delta_l^j$ will be  removed from $\Delta_l$, which effectively reduces the rank of $\Delta_l$ by one. In other words, the rank of $\Delta_l$ is equivalent to the number of non-zero values in $\{\alpha_l^j\}_{j=1}^{k_l}$. 
By learning these selection variables based on their fitness to data, we can automatically determine the rank of $\Delta_l$.   We add a constraint  that the sum of $\{\alpha_l^j\}_{j=1}^{k_l}$ is equal to one: $\sum_{j=1}^{k_l}\alpha_l^j=1$. This constraint renders the optimization of $\{\alpha_l^j\}_{j=1}^{k_l}$ difficult. To address this problem, instead of optimizing $\{\alpha_l^j\}_{j=1}^{k_l}$ directly, we parameterize them using softmax:
\begin{equation}
    \alpha_l^j=\frac{\exp(\beta_l^j)}{\sum_{i=1}^{k_l}\exp(\beta_l^i)},
\end{equation}
and learn the unconstrained variables $\{\beta_l^j\}_{j=1}^{k_l}$.



\subsection{Learn Selection Variables}
\label{sec:meta}
Let $A = \{\alpha^j_l|1 \leq j \leq k_l, 1 \leq l \leq M\}$ denote all selection variables, where $M$ is the number of layers in the pretrained model. 
We propose a meta learning based approach to learn $A$. 
Let $\mathcal{L}_{tr}$ denote the downstream task's training loss defined on a training dataset $D_{tr}$. 
Given the weight parameters $W_l=\widetilde{W}_l+\Delta_l$ at layer $l$ in the downstream model, we first perform a one-step gradient descent update of $W_l$:
\begin{equation}
\label{eq:low}
   \widehat{W}_l= W_l-\eta \nabla_{W_l}\mathcal{L}_{tr}(\{W_l\}_{l=1}^M,D_{tr}), 
\end{equation}
where $\eta$ is a learning rate. Then we evaluate $\{\widehat{W}_l\}_{l=1}^M$ on a validation dataset $D_{val}$. The validation loss $\mathcal{L}_{val}(\{\widehat{W}_l\}_{l=1}^M,D_{val})$ is a function of $A$ since $\mathcal{L}_{val}$ depends on $\{\widehat{W}_l\}_{l=1}^M$ which depends on $A$. 
We optimize $A$ by minimizing the validation loss:
\begin{equation}
\label{eq:meta}
\textrm{min}_A\; \mathcal{L}_{val}(\{\widehat{W}_l\}_{l=1}^M,D_{val}).
\end{equation}
We use an approximate gradient-based algorithm~\cite{betty} to solve this problem. The updates  of $W$ and $A$ in Eq.(\ref{eq:low}) and Eq.(\ref{eq:meta}) are iteratively performed until convergence.

\subsection{Determine Matrix Rank} 
\label{sec:rank}
Given the optimally learned  selection variables $A^*$, we determine the rank of each update matrix based on $A^*$. For each layer $l$, we count the number of entries in  $\{\alpha_l^j\}_{j=1}^{k_l}$ that satisfy $\alpha_l^j\geq\lambda$, where $\lambda$ denotes a threshold. This number would be the optimal rank  for $\Delta_l$. We set $\lambda$ to be $\lambda=1/k_l$. This threshold guarantees the automatically determined rank is at least one.

\subsection{Retrain Update Matrices}
The thresholding operations in Section~\ref{sec:rank} incurs a discrepancy: when training the update matrices in Section~\ref{sec:meta}, all rank-1 matrices are used to make predictions; however, after thresholding, some rank-1 matrices are dropped, which may hurt performance. To bridge this discrepancy, we retrain the update matrices. Specifically, for each update matrix, we set its rank to be the optimal value determined in Section~\ref{sec:rank}, then train them by minimizing the finetuning loss on the combination of training and validation datasets.

\section{Experiments}

\begin{table*}[t]
\centering
\begin{tabular}{l|c|ccccccccc}
\toprule
{Method}&Params & {CoLA} & {SST-2} &{MRPC} & QQP &MNLI &QNLI &RTE&STS-B&Avg.\\
\midrule
Full FT &125.0M&61.6 &94.8 &89.3& 90.3&86.7&92.8&76.9&91.2&85.5\\
\midrule
Adapter&0.9M&58.8&94.0&88.4&89.1&86.5&92.5&71.2&89.9&83.8\\
LoRA  &0.3M& 59.0 & 94.5 &89.1&89.6 &86.9&92.9&75.8&\bf 91.1&84.9\\
AdaLoRA &0.3M& 58.8 & 94.0 &\bf 89.4&89.9&\bf87.0&\bf93.0&75.9&90.6&85.0\\
AutoLoRA &0.3M&\bf61.3 &\bf94.9&\bf89.4&\bf 90.3&\bf87.0&92.9&\bf77.0&90.8&\bf 85.5\\
\bottomrule
\end{tabular}
\caption{\label{glue}
Performance and the number of parameters (Params) of AutoLoRA and baseline methods for finetuning RoBERTa-base on the GLUE benchmark. Higher value is better for all metrics. The best results  are shown in \textbf{bold}. We also provide the performance of full finetuning (Full FT) for reference.}
\end{table*}



\subsection{Experimental Setup}

The baseline methods used in this work include  Adapter \cite{adapter}, LoRA \cite{lora},  and AdaLoRA \cite{adalora}.

We examine the efficacy of AutoLoRA by finetuning a RoBERTa-base model \cite{RoBERTa}, a RoBERTa-large model, and a GPT2-medium model \cite{gpt2} on natural language understanding (NLU), natural language generation (NLG),  and sequence labeling datasets. We include detailed comparison of these two pretrained models in Appendix \ref{sec:appendix_model}.

\begin{figure}[t]
    \centering
    \includegraphics[width=0.48\textwidth]{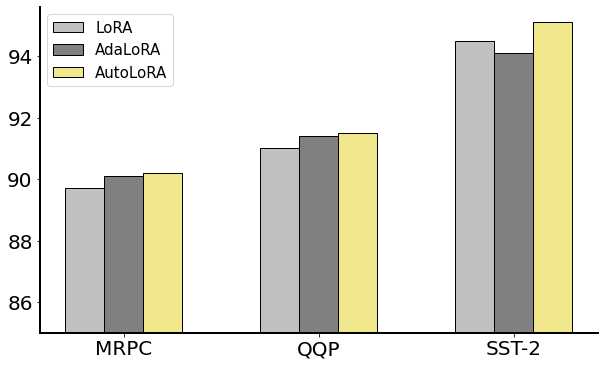}
    \caption{Results of finetuning the RoBERTa-large model on the MRPC, QQP, and SST-2 datasets. Y-axis represents accuracy on GLUE development sets.}
    \label{fig:2}
\end{figure}

A Transformer~\cite{transformer} model consists of several stacked Transformer blocks (layers), and each block contains a multi-head attention (MHA) module and a fully-connected neural network. Each head in an MHA module includes a query projection layer, a key projection layer, and a value projection layer. In adherence to the standard setting in LoRA, we select only the query and value projection layers  as trainable LoRA layers, leaving other layers frozen. Both RoBERTa-base and GPT2-medium possess 12 Transformer layers, which results in  24 trainable LoRA layers. The RoBERTa-large model, with  24 Transformer layers, has 48 trainable LoRA layers.

We set the initial dimension of selection variables $\alpha_l$ to be 8 at each layer, i.e., $k_l=8$. The rank for each layer in LoRA baselines is set as 4, resulting in a similar number of trainable parameters as that in AutoLoRA. We use AdamW~\cite{adamw} as the optimizer for both AutoLoRA and baseline methods. We set the batch size as 16 for NLU and NLG tasks, and 32 for the  sequence labeling task. We set the learning rate for optimizing weight parameters $W$ in Eq.(\ref{eq:low}) to be $1e-4$, and the learning rate for optimizing selection variables $A$ in Eq.(\ref{eq:meta}) to be $1e-3$. All experiments were conducted on NVIDIA A100 GPUs. Our implementation is based on Pytorch \cite{pytorch}, HuggingFace Transformers \cite{huggingface}, and the Betty library \cite{betty}.

\subsection{Experiments on Natural Language Understanding Tasks} 
\label{sec:nlu}

We conduct extensive experiments on eight datasets from the General Language Understanding Evaluation (GLUE) benchmark \cite{glue} to evaluate the performance of AutoLoRA on NLU tasks. The 
GLUE benchmark  contains single sentence classification, sentence pair classification, and regression tasks for  language acceptability evaluation,  sentiment analysis, sentence similarity measurement, and natural language inference. 
We use accuracy as the evaluation metrics for the SST-2, MRPC, QQP, MNLI, QNLI, and RTE tasks. We use  Matthew's correlation for the CoLA  task and Spearman's correlation for the STS-B task. 
Since the test sets of the GLUE benchmark are  not publicly available, following previous studies \cite{subnetwork}, we use the AutoLoRA framework to finetune a RoBERTa-base model on the GLUE training set and evaluate it on the GLUE development set. 
We split the original training set into a new training set and a validation set with a ratio of 1:1, which are used as $D_{tr}$ and $D_{val}$ in Eq.(\ref{eq:low}) and Eq.(\ref{eq:meta}) respectively. Please note that baselines methods are trained on the original training set and our method does not unfairly use more data than baselines.

\begin{table*}[t]
\centering
\begin{tabular}{l|c|ccccc|ccc}
\toprule
& &\multicolumn{5}{c}{E2E} & \multicolumn{3}{|c}{WebNLG}\\
Method&Param&BLEU& NIST&MET&ROUGE-L&CIDEr&BLEU&MET&TER\\
\midrule
Full FT&354.9M&68.0&8.61&46.1&69.0&2.38&46.5&38.0&0.53\\
\midrule
Adapter&11.1M&67.0&8.50&45.2&66.9&2.31&50.2&38.0&0.46\\
LoRA&0.3M & 67.1&8.54&45.7&68.0&2.33& 50.7&39.5&0.46\\
AdaLoRA&0.3M & 67.0 & 8.55 & 45.5 & 68.1 &2.32 & 50.6&39.4 &\bf0.44\\
AutoLoRA &0.3M&\bf67.9&\bf8.68&\bf46.0&\bf68.9&\bf2.37&\bf50.8&\bf39.6&\bf0.44\\
\bottomrule
\end{tabular}
\caption{\label{nlg}
Performance and the number of parameters of AutoLoRA and baseline methods for finetuning GPT-medium  
on the E2E and WebNLG datasets. Higher value is better for all metrics except TER.  Lower TER value indicates better performance.}
\vspace{-0.4cm}
\end{table*}

Table \ref{glue} shows the performance of AutoLoRA on the GLUE development sets, compared with  baseline methods. AutoLoRA achieves the  best performance on 6 out of 8 datasets, and obtains an average performance of 85.5, outperforming all baseline methods. As AutoLoRA outperforms LoRA  on average, we can conclude that the optimal ranks learned by AutoLoRA are better than the manually tuned ranks in LoRA. The reasons are two-fold. First, AutoLoRA allows different layers to have distinct ranks, sufficiently accounting for the fact that different layers have varying properties and need to have layer-specific amounts of tunable parameters. In contrast, LoRA uniformly uses the same rank for all layers, without considering the difference across layers. Second, AutoLoRA  learns the continuous selection variables (which determine the ranks) by maximizing the finetuning performance on validation data via gradient descent. The search space is continuous, which allows more comprehensive exploration of  rank configurations. 
In contrast, LoRA performs manual tuning of ranks in a discrete space, where the number of rank configurations is relatively limited.

Furthermore, AutoLoRA  outperforms the AdaLoRA baseline on average. The reason is that AdaLoRA uses a single dataset to simultaneously learn rank-1 matrices and their importance scores, which can easily lead to overfitting. In contrast, our method splits the training dataset into two disjoint sets, learns rank-1 matrices on  one set, and optimizes selection  variables on the other set, which is more resilient to overfitting. 

In addition, we present the results of fully finetuning a RoBERTa-base model. Results indicate that AutoLoRA attains performance on par with the full finetuning method, while utilizing significantly fewer parameters.

We further examine the efficacy of AutoLoRA with larger pretrained models. Specifically, we applied  AutoLoRA to finetune a RoBERTa-large model \cite{RoBERTa} on the  MRPC, QQP, and SST-2 datasets. The RoBERTa-large model comprises 355 million parameters, in contrast to the RoBERTa-base, which only contains 125 million. As shown in Figure \ref{fig:2}, the performance of AutoLoRA surpasses both baseline methods across all three datasets, demonstrating AutoLoRA's robust effectiveness in  finetuning  pretrained models with various sizes.  


\subsection{Experiments on Natural Language Generation Tasks}

In addition to NLU tasks, we also evaluate the effectiveness of AutoLoRA in NLG tasks. The experiments were conducted on two datasets:  E2E \cite{e2e} and WebNLG \cite{webnlg}. The E2E dataset contains around 50,000 data-sentence pairs in the restaurant domain. Given the data record of a restaurant, the task is to generate a text description for the restaurant. The WebNLG dataset contains more than 10,000 data-sentence pairs extracted from DBpedia. The data contains triples with a format of (subject, property, object), and the task is to generate a  text as a  verbalisation of these triples. We use BLEU~\cite{bleu}, NIST~\cite{nist}, METEOR~\cite{meteor}, ROUGE-L~\cite{rouge},  and CIDEr~\cite{cider} as evaluation metrics for the E2E dataset. For the WebNLG dataset, we use BLEU, METEOR, and TER~\cite{ter} as evaluation metrics. AutoLoRA was applied to finetune a GPT-medium model.

\begin{table}
\centering
\begin{tabular}{l|c|cc}
\toprule
Method&Param&Precision&F1\\
\midrule
Full FT&125.0M &70.3&74.9\\
\midrule
Adapter&0.9M &66.9&71.3\\
LoRA &0.3M&68.5& 72.2\\
AdaLoRA &0.3M& 69.4&73.0\\
AutoLoRA &0.3M&\bf 70.1&\bf74.2\\
\bottomrule
\end{tabular}
\caption{\label{bionlp}
Performance and the number of parameters of AutoLoRA and baseline methods for finetuning a RoBERTa-base model on the BioNLP dataset.} 
\end{table}

\begin{table*}[t]
    \centering
    \begin{tabular}{l|ccccccccc}
    \toprule
    {Method} & {CoLA} & {SST-2} &{MRPC} & QQP &MNLI &QNLI &RTE&STS-B&Avg.\\
    \midrule
         AutoLoRA (w/o cst.)&61.0&93.7& 88.5 &90.0&\bf87.2&92.1& \bf77.5&90.5&85.1 \\
         AutoLoRA (sigmoid)&59.7&94.1& 88.3 &89.8&86.9&92.6&75.7&90.7& 84.7\\
         AutoLoRA ($\eta=0$)&61.2&94.8&89.3&90.1&87.1&92.8& 77.3&90.5&85.2\\
         AutoLoRA &\bf61.3 &\bf94.9&\bf89.4& \bf90.3&87.0&\bf92.9&77.0&\bf90.8& \bf85.5\\
    \bottomrule
    \end{tabular}
    \caption{Ablation studies. Performance comparison of AutoLoRA, AutoLoRA without constraint (AutoLoRA w/o cst.),  AutoLoRA with element-wise sigmoid (AutoLoRA sigmoid), and AutoLoRA with $\eta=0$. We evaluated all methods on the development sets of the GLUE benchmark.}
    \vspace{-0.2cm}
    \label{tab:abl}
\end{table*}


Table \ref{nlg} shows the performance of AutoLoRA on the E2E test set and  WebNLG test set. AutoLoRA achieves the best performance in terms of all five metrics on the E2E dataset. It outperforms or is on par with baseline methods on the WebNLG dataset in terms of all three metrics. This demonstrates  AutoLoRA's effectiveness in finetuning pretrained models for NLG tasks. The analysis of reasons that  AutoLoRA outperforms LoRA and AdaLoRA is similar to that in Section~\ref{sec:nlu}. 
Moreover, the performance of AutoLoRA is on par with that of the full finetuning method, while the number of parameters in AutoLoRA is substantially less. 


\subsection{Experiments on Sequence Labeling}

In this section, we evaluate AutoLoRA on a sequence labeling task. Different from the GLUE tasks which perform classification on an entire sentence (focusing on capturing  global semantics), sequence labeling performs classification on each token in a sentence (emphasizing capturing local context). 
The experiments were conducted on the  BioNLP \cite{bionlp} dataset, which is a Named Entity Recognition dataset  containing  biological entities such as DNA, RNA, and protein. F1 is used as the evaluation metric. AutoLoRA was applied to finetune a RoBERTa-base model for this task.

Table \ref{bionlp} shows the performance of AutoLoRA on the BioNLP test set, compared with baseline methods.  AutoLoRA outperforms all baseline methods in terms of F1 score. The analysis of reasons is similar to that in Section~\ref{sec:nlu}. In line with our previous findings on NLU and NLG tasks, AutoLoRA can effectively finetune pretrained models for sequence labeling.

\begin{figure*}
    \centering
    \includegraphics[width=0.8\textwidth]{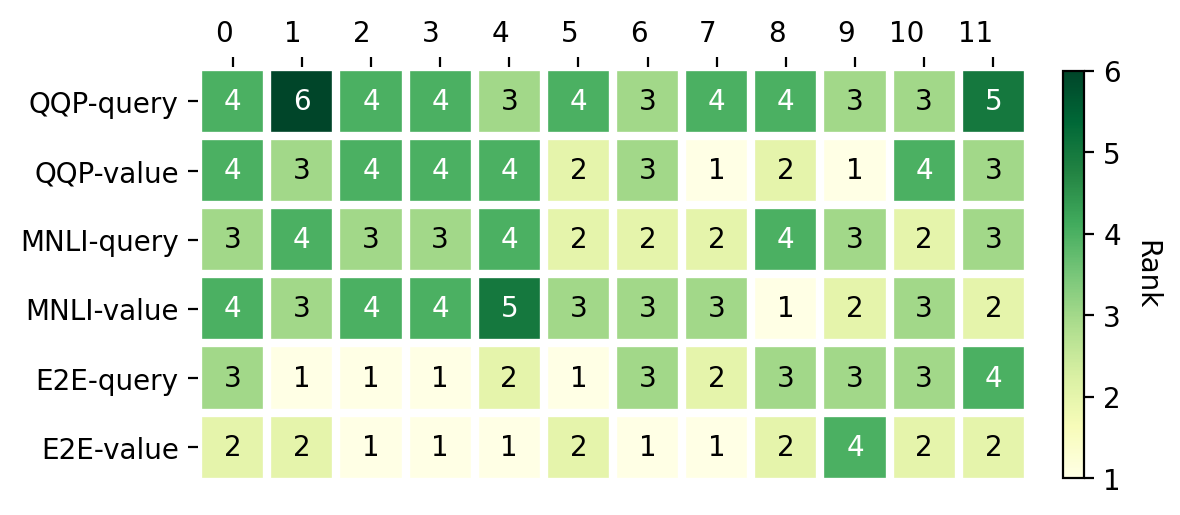}
    \caption{Optimal ranks of the LoRA layers  obtained by AutoLoRA on the QQP, MNLI, and E2E datasets. We finetuned  RoBERTa-base  on  QQP and MNLI, and  GPT2-medium  on  E2E. Both RoBERTa-base and GPT2-medium  consist of 12 Transformer layers. We only search for the ranks in the query and value projection layers.}
    \label{fig:3}
    \vspace{-0.3cm}
\end{figure*}

\subsection{Ablation Studies}
In this section, we perform ablation studies to investigate the effectiveness of individual modules in our method. The studies were performed on the GLUE benchmark.

\paragraph{No Constraints.}  We examine the effectiveness of the sum-to-one constraint in Section \ref{sec:rep} by removing the constraints from AutoLoRA.  Specifically, we directly use a threshold $\lambda=\sum_{j=1}^{k_l}\alpha_l^j/{k_l}$ to obtain the optimal discrete ranks without any constraints in the meta learning process (AutoLoRA w/o cst.). 
Results in Table \ref{tab:abl} show that AutoLoRA outperforms this ablation setting on average, indicating the effectiveness of this sum-to-one constraint. The reason is that adding such a constraint can make the selection variables better represent the relative importance of rank-1 matrices, which facilitate accurate pruning of less important rank-1 matrices.

\paragraph{Element-wise Sigmoid.} We further examine the effectiveness of the sum-to-one  constraint in Section \ref{sec:rep} by comparing  
AutoLoRA with an ablation setting that applies element-wise sigmoid operations on selection variables. 
Specifically, for each $\alpha^j_l$, we use sigmoid to constrain its value into $[0,1]$  in the meta learning process, and a threshold of 0.5 is used to obtain discrete ranks (AutoLoRA sigmoid). Results in Table \ref{tab:abl} show that AutoLoRA outperforms this ablation setting  on average. In this ablation setting, $\alpha^j_l$ no longer directly indicates the relative importance of rank-1 matrices, making it challenging to select an appropriate threshold.

\paragraph{Meta Learning.} We examine the effectiveness of the meta learning framework by setting $\eta=0$ in Algorithm \ref{alg:1}. This ablation setting can be interpreted as an alternative learning method where two optimization steps are carried out alternatively on two different splits of the training dataset. Results in Table \ref{tab:abl} show that AutoLoRA outperforms AutoLoRA ($\eta=0$) on average, demonstrating the efficacy of the meta learning strategy.

\subsection{Qualitative Analysis}
Figure \ref{fig:3} presents the optimal rank determined by AutoLoRA for the QQP, MNLI, and E2E datasets. For the QQP and MNLI datasets, we utilized a RoBERTa-base backbone, while a GPT2-medium backbone was employed for the E2E dataset. In this figure,  column  $i$ corresponds to the $i$-th Transformer block in the pretrained model. Each row  corresponds to a dataset and a  layer type  (query projection and value projection layer).
As can be seen, the optimal ranks learned by AutoLoRA for different layers have varying values. This is aligned with the hypothesis discussed in Section~\ref{sec:intro} that different layers need different matrix ranks. Vanilla LoRA ignores this difference and uniformly uses  the same rank across layers, which leads to inferior performance. Our method provides a computationally efficient mechanism to learn these layer-specific ranks, which takes much less time than grid search (as shown in Section~\ref{sec:cost}).   

\subsection{Computation Costs}
\label{sec:cost}

\begin{table}
\centering
\small
\begin{tabular}{l|c|c|c}
\toprule
Method&AdaLoRA&LoRA+Grid Search &AutoLoRA\\
\midrule
Cost & x1 & x14.29 & x1.91 \\
\bottomrule
\end{tabular}
\caption{\label{tab:cost}
Comparison of average training cost between AutoLoRA and baseline methods on the SST-2, MNLI, and QQP datasets. We normalize the average training time of AdaLoRA as 1.} 
\vspace{-0.4cm}
\end{table}

Table \ref{tab:cost} shows the average training cost of AutoLoRA and two baseline methods on the SST-2, MNLI, and QQP datasets. We normalize the average training time of AdaLoRA  as 1 for reference. In LoRA, we use grid search to tune the ranks, with 16 configurations.  
As can be seen, our method is much more efficient than performing grid search of ranks in LoRA. Grid search is conducted in a discrete space. For each configuration of ranks, LoRA needs to run from scratch, which is very time-consuming. In contrast, our method performs the search in a continuous space via gradient method, which can efficiently explore many configurations without restarting. Compared with AdaLoRA, our method has significantly better performance as shown in Tables~\ref{glue}, \ref{nlg}, and \ref{bionlp}, without substantially increasing the computation costs.  


%

\section{Conclusions and Future Work}

In this paper, we introduce AutoLoRA, a  meta learning based  framework designed to automatically search for the optimal ranks for LoRA layers. Our method associates each rank-1 matrix in LoRA updates with a selection variable and formulates the rank-tuning problem as optimizing the selection variables via meta learning. Thresholding is applied to derive discrete rank values from continuous selection variables and retraining is performed to bridge the gap incurred by thresholding.  Comprehensive experiments show the efficacy of AutoLoRA across various NLP tasks.

Similar to the LoRA method, the LoRA layers in AutoLoRA are manually specified, which may be suboptimal. As a future work, we will investigate how to automatically select LoRA layers, by developing a meta learning framework similar to that in Eq.(\ref{eq:meta}).

\section{Limitations}
In comparison to other rank search techniques like AdaLoRA, our method does introduce some additional  computational and memory overhead. However, as shown in Table \ref{tab:cost}, the increase of training cost is relatively modest. Another limitation is that we did not evaluate  our method on more recent large language models (LLMs), such as LLaMA \cite{llama} and LLaMA-2 \cite{llama2}. It is promising to apply AutoLoRA on these LLMs as they are more powerful compared with previous ones. We did not evaluate our method on LLMs pretrained on non-English texts either. We aim to address these limitations in our future research.

\label{sec:bibtex}


\bibliography{acl_latex}
\bibliographystyle{acl_natbib}
 
\newpage

\appendix
\section{Datasets}
\label{sec:appendix_data}

\begin{table*}[hbt!]
    \centering
    \small
    \begin{tabular}{l|c|c|c|c|c|c|c|c|c}
    \toprule
       & CoLA & RTE & QNLI & STS-B  & MRPC & WNLI & SST-2 & \makecell{MNLI\\(m/mm)} & QQP \\
        \midrule
Train & 8551 & 2490 & 104743 & 5749 & 3668 & 635 & 67349 & 392702 & 363871 \\
Dev & 1043&  277 & 5463 & 1500 & 408 & 71 & 872 & 9815/9832 & 40432 \\
    \bottomrule
    \end{tabular}
    \caption{GLUE dataset statistics.}
    \label{tab:data-stats}
\end{table*}

Table~\ref{tab:data-stats} shows the statistics of the GLUE datasets.


\section{Pretrained Models}

\label{sec:appendix_model}
RoBERTa pretrains a Transformer encoder, which is the same as that in BERT~\cite{bert}. The GPT2 model pretrains a Transformer decoder.  The RoBERTa model is pretrained via  masked token prediction.  The GPT2 model is pretrained via language modeling. 
 RoBERTa  is commonly used for natural language understanding (NLU) tasks while GPT2 is often used for natural language generation (NLG) tasks. 

\section{Hyperparameter Optimization}
Adequate hyperparameter configuration is crucial for machine learning algorithms to achieve top performance. Compared with grid search and simple random search, Bayesian Optimization (BO) \cite{smac3} and gradient-based hyperparameter optimization \cite{gradhyper} have been widely used because of their sample efficiency. For example, SMAC \cite{smac} builds a probabilistic model to estimate the performance of different hyperparameter configurations. It sequentially chooses the next set of hyperparameters to evaluate, with an predefined acquisition function to balance exploration with exploitation in the hyperparameter space. SMAC3 \cite{smac3} improves SMAC by evaluating less promising hyperparameters configurations with fewer instances. c-TPE \cite{ctpe} proposes a constrained tree-structured Parzen estimator to handle constraints such as memory consumption and inference latency of a configuration of hyperparameters. PED-ANOVA \cite{pedanova} highlights the role of good hyperparameter search space in hyperparameter optimization. It derives a algorithm to compute hyperparameter importance with Pearson divergence. On the other hand, \citet{gradhyper} computes the gradients with respect to hyperparameters, and proposes an efficient method to store related  information.

\end{document}